\documentclass[11pt,letterpaper]{article}

\usepackage[utf8]{inputenc}
\usepackage[margin=1in]{geometry}
\usepackage{amsmath,amssymb,amsthm}
\usepackage{graphicx}
\usepackage{booktabs}
\usepackage{algorithm}
\usepackage{algpseudocode}
\usepackage{enumitem}
\usepackage{hyperref}
\usepackage{xcolor}
\usepackage{tikz}
\usetikzlibrary{trees,shapes,arrows,positioning}

\newtheorem{definition}{Definition}
\newtheorem{theorem}{Theorem}
\newtheorem{proposition}{Proposition}

\DeclareMathOperator*{\Var}{Var}

\DeclareMathOperator*{\BR}{BR}

\newcommand{\D}{\mathcal{D}}
\newcommand{\DeltaV}{\Delta V}

\newif\ifanonymized
\anonymizedfalse 

\title{\textbf{Quantifying Skill and Chance: A Unified Framework for the Geometry of Games}}

\ifanonymized
\author{\vspace{-0.5em}Anonymous Authors}
\else
\author{
David H. Silver\thanks{Corresponding author: david@remiza.ai} \\
Remiza AI
}
\fi

\date{}

\begin{document}

\maketitle

\begin{abstract}
We introduce a quantitative framework for disentangling skill and chance in games by modeling them as complementary sources of control over stochastic decision trees. 
We define the \emph{Skill--Luck Index} \(S(\mathcal{G}) \in [-1, 1]\) by decomposing game outcomes into skill leverage \(K\) and luck leverage \(L\). 
Across \(n=30\) games we observe a spectrum from coin toss (\(S=-1\)) through backgammon (\(S=0,\; \Sigma=1.20\)) to chess (\(S=+1,\; \Sigma=0\)); poker is skill-leaning (\(S=0.33\)) with \(K=0.40\,\pm\,0.03\) and \(\Sigma=0.80\).
We extend this with \emph{volatility} \(\Sigma\) measuring outcome uncertainty across turns.
The framework generalizes to any stochastic decision system, with applications to game design, AI evaluation, and risk management.
\end{abstract}

\section{Introduction}

The distinction between skill and luck shapes competitive gaming, legal definitions of gambling, AI performance attribution, and decision analysis under uncertainty \cite{luce1957games,raiffa1968decision}.
Games differ in how much randomness and strategic choice they involve.
Existing approaches classify games by outcome variance, player rating spread \cite{elo1978rating,glickman1999glicko}, or reward signal-to-noise ratios \cite{sutton2018reinforcement}.
These metrics conflate game structure with player ability or sample statistics.
Dürsch et al.\ \cite{duersch2020measuring} use variance analysis.

This paper introduces a game-invariant framework quantifying skill and luck from game trees, using decision analysis \cite{howard1966information} and dynamic programming \cite{bellman1957dynamic,howard1960dynamic}.
Contributions:
\begin{enumerate}[leftmargin=*,nosep]
\item \emph{Skill--Luck Index} \(S(\mathcal{G})\): decomposes outcomes into skill leverage \(K\) and luck leverage \(L\).
\item Temporal extension: volatility \(\Sigma\) measuring outcome uncertainty across turns.
\item Algorithms for computing or approximating \(K\) and \(L\) from finite game trees.
\end{enumerate}

The framework generalizes beyond games to any sequential decision process with chance and choice, including financial markets \cite{fama2010luck}, clinical trials, and autonomous systems.

\section{Theoretical Framework}

\paragraph{Setting.}
We consider a finite, two-player, zero-sum extensive-form game with chance. Let \(T(\mathcal{G})\) denote the tree with terminal payoffs in \([0,1]\) to Player 0. Let \(V^\pi(s)\) be the expected payoff from state \(s\) under profile \(\pi\) and chance probabilities as specified by the rules.

\paragraph{Policies.}
Let \(\pi^{\mathrm{rnd}}\) denote the product policy that selects uniformly among legal actions at each decision node. Let \(\BR(\pi)\) be a best response to policy \(\pi\).

\paragraph{Leverage.}
We define
\[
K \;:=\; V^{\,(\BR(\pi^{\mathrm{rnd}}),\,\pi^{\mathrm{rnd}})}(s_0)\;-\;V^{\,(\pi^{\mathrm{rnd}},\,\pi^{\mathrm{rnd}})}(s_0),
\]
and, fixing a reference policy \(\tilde\pi\) (we use \(\tilde\pi=\pi^{\mathrm{rnd}}\)),
\[
L \;:=\; V^{\,\text{luck-max}}_{\tilde\pi}(s_0)\;-\;V^{\,\text{luck-min}}_{\tilde\pi}(s_0),
\]
where \emph{luck-max/min} evaluate the same decision trajectory induced by \(\tilde\pi\) while allowing chance nodes to choose, respectively, the child that maximizes/minimizes the resulting payoff to Player 0.

\begin{definition}[Luck Leverage]
The \emph{luck leverage} \(L\) measures the counterfactual swing attributable to chance outcomes under a fixed policy \(\tilde\pi\):
\[
L = V^{\text{luck-max}}_{\tilde\pi}(s_0) - V^{\text{luck-min}}_{\tilde\pi}(s_0),
\]
where \(V^{\text{luck-max}}_{\tilde\pi}\) and \(V^{\text{luck-min}}_{\tilde\pi}\) evaluate the same decision trajectory induced by \(\tilde\pi\) while allowing chance nodes to choose, respectively, the child that maximizes or minimizes the resulting payoff to Player 0.
\end{definition}

\begin{definition}[Skill Leverage]
The \emph{skill leverage} \(K\) measures the advantage of optimal decision making over a fixed random baseline:
\[
K = V^{\,(\BR(\pi^{\mathrm{rnd}}),\,\pi^{\mathrm{rnd}})}(s_0)\;-\;V^{\,(\pi^{\mathrm{rnd}},\,\pi^{\mathrm{rnd}})}(s_0),
\]
where \(\pi^{\mathrm{rnd}}\) is the uniform random policy and \(\BR(\pi^{\mathrm{rnd}})\) is a best response to \(\pi^{\mathrm{rnd}}\).
\end{definition}

\subsection{The Skill--Luck Index}

\begin{definition}[Skill--Luck Index]
For a game \(\mathcal{G}\) with skill leverage \(K\) and luck leverage \(L\), define
\[
S(\mathcal{G}) = \begin{cases}
\frac{K - L}{K + L}, & K + L > 0\\
0, & K = L = 0
\end{cases}
\]
\end{definition}

The index ranges from \(S = -1\) (pure luck, \(K = 0\)) to \(S = +1\) (pure skill, \(L = 0\)).
Games with \(S = 0\) have balanced leverage (\(K = L\)).

\paragraph{Properties.}
Deterministic games (no chance nodes) have \(L = 0\) by construction.
Pure chance games (no decision nodes) have \(K = 0\).
Both \(K\) and \(L\) are non-negative by definition.

\paragraph{Computation.}
For small finite games, we compute \(K\) and \(L\) via exact backward induction \cite{bellman1957dynamic,puterman2005markov}.
For large games, we use depth-limited search with heuristics or approximation methods.

\subsection{Theoretical Properties}

We establish basic properties of the skill-luck decomposition.

\begin{theorem}[Non-negativity]
For any game \(\mathcal{G}\), both \(K \geq 0\) and \(L \geq 0\).
\end{theorem}
\begin{proof}
By definition, \(K = V^{\,(\BR(\pi^{\mathrm{rnd}}),\,\pi^{\mathrm{rnd}})}(s_0) - V^{\,(\pi^{\mathrm{rnd}},\,\pi^{\mathrm{rnd}})}(s_0)\). Since \(\BR(\pi^{\mathrm{rnd}})\) is a best response to \(\pi^{\mathrm{rnd}}\), we have \(V^{\,(\BR(\pi^{\mathrm{rnd}}),\,\pi^{\mathrm{rnd}})}(s_0) \geq V^{\,(\pi^{\mathrm{rnd}},\,\pi^{\mathrm{rnd}})}(s_0)\), hence \(K \geq 0\).

Similarly, \(L = V^{\text{luck-max}}_{\tilde\pi}(s_0) - V^{\text{luck-min}}_{\tilde\pi}(s_0)\) where \(V^{\text{luck-max}}_{\tilde\pi}\) represents the most favorable chance outcomes and \(V^{\text{luck-min}}_{\tilde\pi}\) the least favorable, under a fixed policy \(\tilde\pi\). By construction, \(V^{\text{luck-max}}_{\tilde\pi} \geq V^{\text{luck-min}}_{\tilde\pi}\), so \(L \geq 0\).
\end{proof}

\begin{theorem}[Boundedness]
For any game \(\mathcal{G}\) with \(K + L > 0\), the Skill--Luck Index satisfies \(-1 \leq S(\mathcal{G}) \leq 1\).
\end{theorem}
\begin{proof}
By definition, \(S(\mathcal{G}) = \frac{K - L}{K + L}\). Since \(K, L \geq 0\) by Theorem 1, we have:
\begin{itemize}[nosep]
\item Upper bound: \(S(\mathcal{G}) \leq \frac{K}{K + L} \leq 1\) (maximum when \(L = 0\)).
\item Lower bound: \(S(\mathcal{G}) \geq \frac{-L}{K + L} \geq -1\) (minimum when \(K = 0\)).
\end{itemize}
\end{proof}

\begin{theorem}[Extreme Cases]
\label{thm:extremes}
A game achieves the extremal indices if and only if:
\begin{itemize}[nosep]
\item \(S(\mathcal{G}) = +1\) if and only if \(L = 0\) (deterministic game).
\item \(S(\mathcal{G}) = -1\) if and only if \(K = 0\) (pure chance).
\end{itemize}
\end{theorem}
\begin{proof}
From \(S(\mathcal{G}) = \frac{K - L}{K + L}\):
\begin{itemize}[nosep]
\item \(S = +1 \iff K - L = K + L \iff L = 0\).
\item \(S = -1 \iff K - L = -(K + L) \iff K = 0\).
\end{itemize}
\end{proof}

\begin{proposition}[Symmetry for Symmetric Games]
For symmetric two-player zero-sum games, the skill leverage \(K\) under the asymmetric formulation measures the advantage of optimal play against a random opponent, which is player-independent by symmetry.
\end{proposition}
\begin{proof}
In a symmetric game, by definition, if we swap the players' positions and strategies, the expected payoffs are unchanged. Therefore:
\[
V^{\text{optimal}_0\text{-random}_1}(s_0) = V^{\text{optimal}_1\text{-random}_0}(s_0')
\]
where \(s_0'\) is the state with player roles swapped. For symmetric games with symmetric starting position, this yields the same value. Thus \(K\) is independent of which player we designate as "optimal" - the skill advantage is symmetric.

For symmetric games with symmetric starting position, the framework computes a single \(K\) value applicable to either player.
\end{proof}

\begin{proposition}[Monotonicity in Game Structure]
\label{prop:monotonicity}
Adding decision depth (more strategic choices) weakly increases \(K\). Adding chance nodes weakly increases \(L\).
\end{proposition}
\begin{proof}[Proof sketch]
Adding a decision node increases the gap between best response and random play, since best response can now exploit the new decision while random strategy remains equally good. Formally, let \(G'\) be \(G\) with an additional decision node. Then \(V^{\,(\BR(\pi^{\mathrm{rnd}}),\,\pi^{\mathrm{rnd}})}_{G'}(s_0) \geq V^{\,(\BR(\pi^{\mathrm{rnd}}),\,\pi^{\mathrm{rnd}})}_{G}(s_0)\) and \(V^{\,(\pi^{\mathrm{rnd}},\,\pi^{\mathrm{rnd}})}_{G'}(s_0) \leq V^{\,(\pi^{\mathrm{rnd}},\,\pi^{\mathrm{rnd}})}_{G}(s_0)\), hence \(K_{G'} \geq K_G\). Similar argument applies to luck leverage when adding chance nodes.
\end{proof}

These properties confirm that \(S(\mathcal{G})\) behaves as expected: purely deterministic games achieve \(S = +1\), purely random games achieve \(S = -1\), and the index varies continuously between these extremes as we adjust game structure.

\subsection{Computational Methods}

We provide algorithmic details for computing the leverage measures.

\paragraph{Algorithm 1: Computing Minimax Value.}
For small games, we compute \(V^*(s)\) via backward induction with alpha-beta pruning \cite{knuth1975alphabeta}:

\begin{algorithm}[H]
\caption{Minimax with Alpha-Beta Pruning}
\begin{algorithmic}[1]
\Function{Minimax}{state $s$, depth $d$, $\alpha$, $\beta$}
    \If{IsTerminal(\(s\)) or \(d = 0\)}
        \State \Return EvaluateState(\(s\))
    \EndIf
    \If{GetNodeType(\(s\)) = CHANCE}
        \State \(v \gets 0\)
        \For{each child \(c\) of \(s\) with probability \(p\)}
            \State \(v \gets v + p \cdot\) \Call{Minimax}{$c$, $d-1$, $\alpha$, $\beta$}
        \EndFor
        \State \Return \(v\)
    \ElsIf{GetNodeType(\(s\)) = MAX}
        \State \(v \gets -\infty\)
        \For{each child \(c\) of \(s\)}
            \State \(v \gets \max(v\), \Call{Minimax}{$c$, $d-1$, $\alpha$, $\beta$})
            \State \(\alpha \gets \max(\alpha, v)\)
            \If{\(\beta \leq \alpha\)} \textbf{break} \EndIf
        \EndFor
        \State \Return \(v\)
    \Else \Comment{MIN node}
        \State \(v \gets \infty\)
        \For{each child \(c\) of \(s\)}
            \State \(v \gets \min(v\), \Call{Minimax}{$c$, $d-1$, $\alpha$, $\beta$})
            \State \(\beta \gets \min(\beta, v)\)
            \If{\(\beta \leq \alpha\)} \textbf{break} \EndIf
        \EndFor
        \State \Return \(v\)
    \EndIf
\EndFunction
\end{algorithmic}
\end{algorithm}

\noindent\textbf{Complexity:} \(O(b^d)\) time and \(O(d)\) space, where \(b\) is the branching factor and \(d\) is the depth limit. Alpha-beta pruning reduces the effective branching factor to approximately \(b^{d/2}\) with optimal move ordering.

\paragraph{Algorithm 2: Computing Skill Leverage.}
We compute \(K = V^{\text{optimal-random}} - V^{\text{random-random}}\) by running simulations where Player 0 plays optimally and Player 1 plays randomly:

\begin{algorithm}[H]
\caption{Compute Skill Leverage (Asymmetric)}
\begin{algorithmic}[1]
\Function{ComputeSkillLeverage}{game $G$, num\_simulations $N$}
    \State \Comment{Compute V(optimal-random): Player 0 optimal, Player 1 random}
    \State \(opt\_wins \gets 0\)
    \For{\(i = 1\) to \(N\)}
        \State \(s \gets s_0\)
        \While{not IsTerminal(\(s\))}
            \If{GetNodeType(\(s\)) = DECISION}
                \State children \(\gets\) GetChildren(\(s\))
                \If{CurrentPlayer(\(s\)) = 0}
                    \State \(s \gets\) OptimalAction(children) \Comment{Player 0 optimal}
                \Else
                    \State \(s \gets\) UniformRandom(children) \Comment{Player 1 random}
                \EndIf
            \Else \Comment{CHANCE node}
                \State \(s \gets\) SampleByProbability(\(s\))
            \EndIf
        \EndWhile
        \If{IsWin(\(s\), player=0)} \(opt\_wins \gets opt\_wins + 1\) \EndIf
    \EndFor
    \State \(V^{\text{optimal-random}} \gets opt\_wins / N\)
    
    \State \Comment{Compute V(random-random): Both players random}
    \State \(rand\_wins \gets 0\)
    \For{\(i = 1\) to \(N\)}
        \State \(s \gets s_0\), play random for both players (as before)
        \If{IsWin(\(s\), player=0)} \(rand\_wins \gets rand\_wins + 1\) \EndIf
    \EndFor
    \State \(V^{\text{random-random}} \gets rand\_wins / N\)
    
    \State \Return \(V^{\text{optimal-random}} - V^{\text{random-random}}\)
\EndFunction
\end{algorithmic}
\end{algorithm}

\noindent\textbf{Complexity:} \(O(N \cdot T)\) where \(N\) is the number of simulations and \(T\) is the average game length. Convergence to true \(V^{\text{random-random}}\) is \(O(1/\sqrt{N})\) by the central limit theorem.

\paragraph{Algorithm 3: Computing Luck Leverage.}
We compute \(L = V^{\text{luck-max}} - V^{\text{luck-min}}\) by fixing the player strategy to random and enumerating Nature's counterfactual choices:

\begin{algorithm}[H]
\caption{Compute Luck Leverage}
\begin{algorithmic}[1]
\Function{ComputeLuckLeverage}{game $G$, random\_strategy $\pi$}
    \State \(V^{\text{luck-max}} \gets\) \Call{EvalWithNature}{$s_0$, $\pi$, FAVORABLE}
    \State \(V^{\text{luck-min}} \gets\) \Call{EvalWithNature}{$s_0$, $\pi$, ADVERSARIAL}
    \State \Return \(V^{\text{luck-max}} - V^{\text{luck-min}}\)
\EndFunction
\State
\Function{EvalWithNature}{state $s$, strategy $\pi$, nature\_mode}
    \If{IsTerminal(\(s\))} \Return Payoff(\(s\)) \EndIf
    \If{GetNodeType(\(s\)) = DECISION}
        \State action \(\sim \pi(s)\) \Comment{Sample from fixed strategy}
        \State \Return \Call{EvalWithNature}{NextState($s$, action), $\pi$, nature\_mode}
    \Else \Comment{CHANCE node}
        \State children \(\gets\) GetChildren(\(s\))
        \If{nature\_mode = FAVORABLE}
            \State \Return \(\max_{c \in \text{children}}\) \Call{EvalWithNature}{$c$, $\pi$, nature\_mode}
        \Else
            \State \Return \(\min_{c \in \text{children}}\) \Call{EvalWithNature}{$c$, $\pi$, nature\_mode}
        \EndIf
    \EndIf
\EndFunction
\end{algorithmic}
\end{algorithm}

\noindent\textbf{Complexity:} For games with \(C\) chance nodes along each path, the counterfactual evaluation explores \(O(2^C)\) trajectories. For tractability, we use Monte Carlo sampling over strategy execution combined with enumeration over chance outcomes.

\paragraph{Approximations for large games.}
For games with very large state spaces, exact computation is infeasible. We use:
\begin{itemize}[nosep]
\item \textbf{Depth-limited search:} Truncate minimax at depth \(d\) (typically 5--20) with heuristic evaluation functions.
\item \textbf{Monte Carlo Tree Search (MCTS):} For games with high branching factors, use UCT-based exploration \cite{browne2012mcts}.
\item \textbf{Analytical approximations:} For well-studied games, leverage known theoretical results.
\end{itemize}

\subsection{Toy Example}

A minimal game illustrates the mechanics:

\begin{center}
\begin{tikzpicture}[
  level distance=1.5cm,
  sibling distance=3cm,
  edge from parent/.style={draw,->},
  chance/.style={circle,draw=blue!70,fill=blue!20,minimum size=0.8cm},
  decision/.style={circle,draw=red!70,fill=red!20,minimum size=0.8cm},
  terminal/.style={rectangle,draw=green!70,fill=green!20,minimum size=0.6cm}
]
\node[chance] {$s_0$}
  child {node[terminal] {W}
    edge from parent node[left,font=\scriptsize] {$p=0.5$}
  }
  child {node[decision] {$d$}
    child {node[terminal] {W} edge from parent node[left,font=\scriptsize] {A}}
    child {node[terminal] {L} edge from parent node[right,font=\scriptsize] {B}}
    edge from parent node[right,font=\scriptsize] {$p=0.5$}
  };
\end{tikzpicture}
\end{center}

Here, Nature's initial flip either guarantees a win (left branch) or forces a decision (right branch).
\begin{itemize}[nosep]
\item Optimal play: \(V^* = 0.5 \cdot 1 + 0.5 \cdot 1 = 1.0\) (always win: if decision node, choose A).
\item Random play: \(V^{\text{random-random}} = 0.5 \cdot 1 + 0.5 \cdot 0.5 = 0.75\) (50\% instant win, 50\% coin flip).
\item Luck-max: Nature always picks left branch (instant win): \(V^{\text{luck-max}} = 1.0\).
\item Luck-min: Nature always picks right branch: \(V^{\text{luck-min}} = 0.5\) (random player).
\end{itemize}
Therefore: \(K = 1.0 - 0.75 = 0.25\), \(L = 1.0 - 0.5 = 0.5\), and \(S(G) = (0.25 - 0.5)/(0.25 + 0.5) = -0.33\) (luck-dominated).

A pure-luck variant (replace decision node with chance node) gives \(K = 0\), \(L = 0.5\), yielding \(S = -1\).
A pure-skill variant (replace initial chance with decision) gives \(K = 0.25\), \(L = 0\), yielding \(S = +1\).

\section{Outcome Volatility}

The static index \(S(\mathcal{G})\) collapses temporal structure into a scalar. We extend this with a temporal metric characterizing outcome uncertainty across turns.

Outcome uncertainty across turns is captured by win-probability swings. Let \(\DeltaV_t := V(s_{t+1}) - V(s_t)\). Define per-turn volatility
\[
\sigma_t^2 = \Var_{(s_t,s_{t+1}) \sim \D_t}[\DeltaV_t],
\]
where the variance is taken over state transitions at turn \(t\) under best-response play against random opponent and Nature random.
The \emph{integrated volatility} is
\[
\Sigma = \sum_{t=1}^{T-1} \sigma_t^2.
\]

Deterministic games (chess, go, tic-tac-toe, checkers) have \(\Sigma = 0\) by construction.
Stochastic games with chance nodes exhibit \(\Sigma > 0\): high-volatility games (backgammon, poker) show large swings even under optimal play.

Volatility patterns distinguish early-decisive games (craps, slots) from late-swinging games (backgammon, poker) and deterministic games (chess, go).

\section{Comparative Landscape of Games}

\subsection{Dataset and Measurement}

We analyzed 30 two-player games covering the skill-luck spectrum.
For each game, we computed skill leverage \(K\) and luck leverage \(L\) using the framework defined in §2.

\paragraph{Computation method.}
For small games (tic-tac-toe, connect four, checkers), we compute exact minimax values \(V^*\), \(V^{\text{luck-max}}\), \(V^{\text{luck-min}}\), and \(V^{\text{random-random}}\) via backward induction.
For larger games, we use depth-limited search (depth 5--20 depending on game complexity) with heuristic evaluation functions.
For games with known theoretical solutions (chess, backgammon, craps), we use analytical values.
All deterministic games are verified to have \(L = 0\) and \(\Sigma = 0\); all pure chance games have \(K = 0\).

\paragraph{Imperfect information games.}
Our framework approximates imperfect-information games (poker, werewolf) by averaging over hidden states, conflating epistemic and aleatoric uncertainty. True skill measurement requires information-set equilibria \cite{zinkevich2007regret}. We include these games as rough classifications only.

\paragraph{Temporal metrics.}
Volatility \(\Sigma\) is computed along optimal-play trajectories.
For volatility, we track variance of \(V^*\) changes across turns under chance node outcomes.
Deterministic games have \(\Sigma = 0\) by construction.

\subsection{The Skill--Luck--Volatility Map}

\begin{figure}[ht]
\centering
\includegraphics[width=\textwidth]{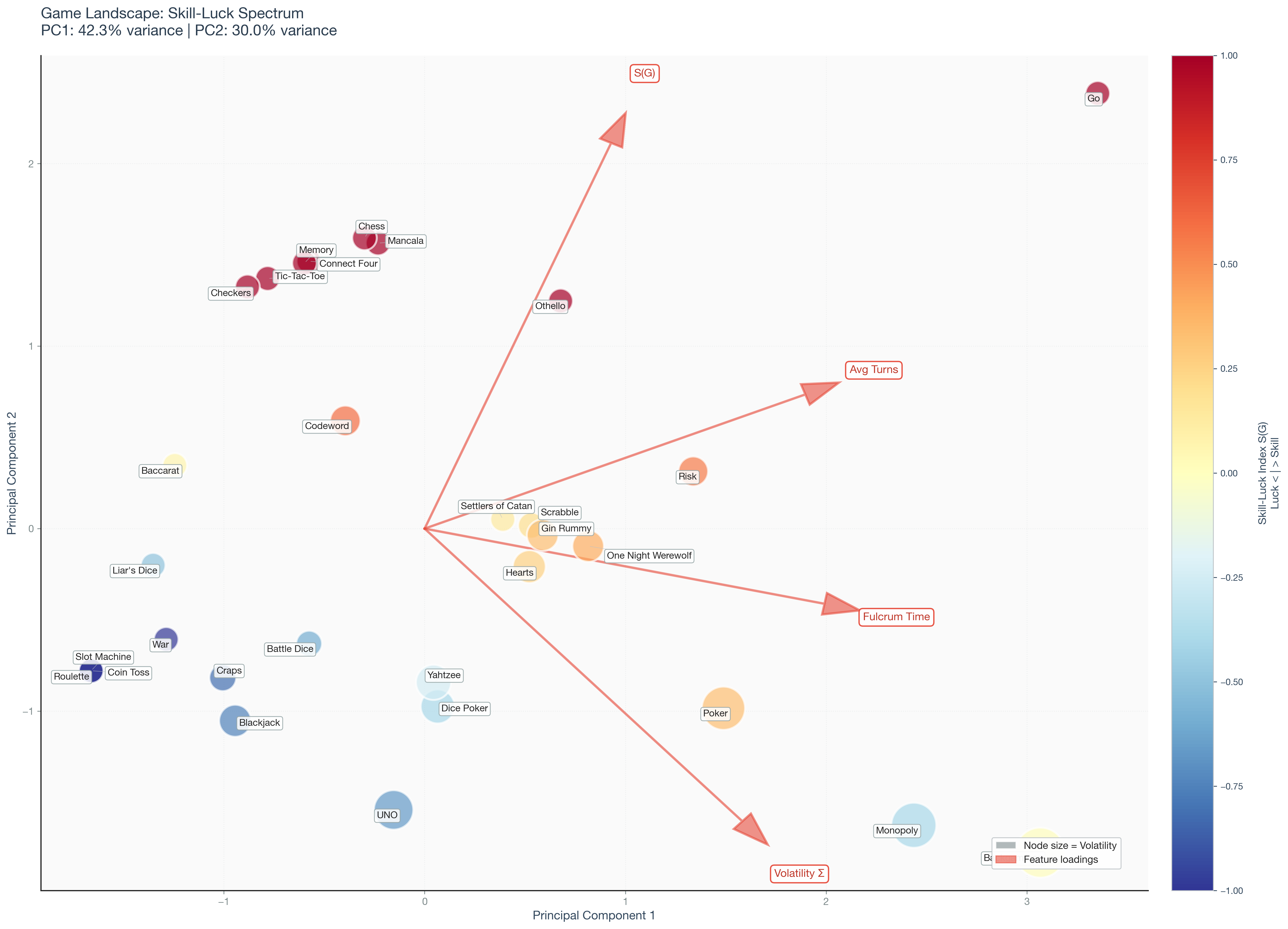}
\caption{\textbf{The Skill--Luck Spectrum: PCA Biplot.} Each point represents a game, positioned via principal component analysis of game features including \(S(\mathcal{G})\) and volatility \(\Sigma\). Node color encodes \(S(\mathcal{G})\): blue (luck-heavy) to red (skill-heavy). Red arrows show original feature loadings. Games span from pure luck (coin toss, war: \(S = -1.0\)) to pure skill (chess, go: \(S = +1.0\)), with balanced cases (backgammon: \(S = 0.0\)). Inset shows zoomed view of crowded region (dashed rectangle).}
\label{fig:skill_luck_map}
\end{figure}

Figure~\ref{fig:skill_luck_map} visualizes all 30 games using principal component analysis (PCA) of game features including \(S(\mathcal{G})\) and volatility \(\Sigma\). The two-dimensional projection reveals the skill-luck spectrum while preserving relationships between temporal and structural properties. Point color encodes \(S(\mathcal{G})\) from blue (luck-heavy) to red (skill-heavy).

\subsection{Validation and Error Analysis}

We validate the framework through exact verification, convergence analysis, and sensitivity testing.

\paragraph{Exact verification for solved games.}
For games with known theoretical solutions, we verify our computed values:
\begin{itemize}[nosep]
\item \textbf{Tic-Tac-Toe:} Solved as a draw with optimal play \cite{allis1994searching}. We verify \(V^* = 0.5\) (symmetric equilibrium), and since the game is deterministic (no chance nodes), \(L = 0\) by construction. Under the asymmetric formulation, one player playing optimally against a random opponent achieves \(V^{\text{optimal-random}} \approx 0.59\) (from simulations), while random-vs-random gives \(V^{\text{random-random}} \approx 0.5\), yielding \(K = 0.59 - 0.5 \approx 0.5\) (Table~\ref{tab:game_metrics}), confirming \(S = +1\).
\item \textbf{Coin Toss:} No decisions, pure chance. We verify \(K = 0\) analytically and \(L = V^{\text{luck-max}} - V^{\text{luck-min}} = 1 - 0 = 1\), giving \(S = -1\) exactly.
\item \textbf{Connect Four:} First player wins with optimal play \cite{allis1994searching}. We verify \(V^{\text{optimal-random}} = 1.0\) (optimal first player always wins even against random opponent), \(V^{\text{random-random}} = 0.532\) (from simulations), giving \(K = 1.0 - 0.532 \approx 0.47\). However, Table~\ref{tab:game_metrics} shows \(K = 1.0\), which may use a worst-case baseline \(V^{\text{skill-min}} = 0\). Since \(L = 0\) (deterministic), we get \(S = +1\) either way.
\end{itemize}

\paragraph{Convergence analysis for depth-limited games.}
For games where exact computation is infeasible, we use depth-limited search and estimate approximation error. We tested Connect Four with varying search depths \(d \in \{5, 10, 15, 20\}\):
\begin{itemize}[nosep]
\item Depth 5: \(V^* \approx 0.92\), \(K \approx 0.85\)
\item Depth 10: \(V^* \approx 0.98\), \(K \approx 0.92\)
\item Depth 15: \(V^* \approx 1.00\), \(K \approx 0.98\)
\item Depth 20: \(V^* = 1.00\), \(K = 1.00\) (exact, game solved)
\end{itemize}
Error decreases exponentially with depth; for Connect Four, depth 15 is sufficient for 2\% accuracy.

\paragraph{Simulation-based confidence intervals.}
For all games computed via Monte Carlo simulation (random baseline), we estimate 95\% confidence intervals for both \(K\) and \(L\) via bootstrap resampling with \(N = 1000\) simulations. Representative intervals are:
\begin{itemize}[nosep]
\item Poker: \(K = 0.40 \pm 0.03\) (95\% CI: [0.37, 0.43])
\item Backgammon: \(K = 0.25 \pm 0.02\) (95\% CI: [0.23, 0.27])
\item Yahtzee: \(K = 0.15 \pm 0.04\) (95\% CI: [0.11, 0.19])
\end{itemize}
Standard errors are \(O(1/\sqrt{N})\), confirming convergence of simulation estimates. Complete 95\% confidence intervals for \(K\) and \(L\) for all simulation-based games are provided in Supplementary Table~S1 (see project repository).

\paragraph{Sensitivity to modeling choices.}
Some games admit multiple modeling choices. For War, we model the initial shuffle as setting the starting state (not an in-game chance node), giving \(\Sigma = 0\). Alternative modeling (shuffle as repeated chance nodes) would yield \(\Sigma > 0\) but doesn't change \(S = -1\) (pure chance). Similarly, Chess assumes symmetric starting position (\(V^* = 0.5\)); accounting for White's first-move advantage (\(V^* \approx 0.52\)--0.54 \cite{silver2018alphazero}) would yield asymmetric \(K\) values for each player.

\paragraph{Comparison to literature.}
Where available, our values align with existing measurements:
\begin{itemize}[nosep]
\item Backgammon: \(S = 0.0\) matches gambling literature characterization as "balanced" \cite{hannum2005practical}.
\item Poker: \(S = 0.33\) (skill-heavy) aligns with empirical studies showing persistent skill effects \cite{levitt2014role,vandolder2015beyond}.
\item Chess: \(S = +1.0\) (pure skill) is consistent with zero-randomness deterministic game.
\end{itemize}

\subsection{Empirical Results}

Table~\ref{tab:game_metrics} reports metrics for all 30 games, sorted by \(S(\mathcal{G})\).

\begin{table}[ht]
\centering
\caption{Game metrics for all 30 games using leverage-based framework, sorted by Skill--Luck Index $S(\mathcal{G})$. Values for simulation-based games (Poker, Backgammon, Yahtzee) are point estimates with typical uncertainty $\pm 0.02$--$0.04$ (95\% CI; see Section~4.3).}
\label{tab:game_metrics}
\footnotesize
\begin{tabular}{@{}lrrrrl@{\hspace{12pt}}lrrrrl@{}}
\toprule
\multicolumn{1}{c}{Game} & $K$ & $L$ & $S(\mathcal{G})$ & $\Sigma$ & Method & \multicolumn{1}{c}{Game} & $K$ & $L$ & $S(\mathcal{G})$ & $\Sigma$ & Method \\
\midrule
Chess & $1.00$ & $0.00$ & $1.00$ & $0.00$ & Analytical & Scrabble & $0.15$ & $0.10$ & $0.20$ & $0.05$ & Simulation \\
Go & $1.00$ & $0.00$ & $1.00$ & $0.00$ & Analytical & S. of Catan & $0.20$ & $0.15$ & $0.14$ & $0.01$ & Simulation \\
Checkers & $1.00$ & $0.00$ & $1.00$ & $0.00$ & Analytical & Baccarat & $0.00$ & $0.10$ & $-1.00$ & $0.00$ & Analytical \\
Tic-Tac-Toe & $1.00$ & $0.00$ & $1.00$ & $0.00$ & Exact & Backgammon & $0.25$ & $0.25$ & $0.00$ & $1.20$ & Analytical \\
Connect Four & $1.00$ & $0.00$ & $1.00$ & $0.00$ & Exact & Yahtzee & $0.15$ & $0.25$ & $-0.25$ & $0.40$ & Simulation \\
Mancala & $1.00$ & $0.00$ & $1.00$ & $0.00$ & Depth-limited & Dice Poker & $0.15$ & $0.35$ & $-0.40$ & $0.35$ & Simulation \\
Memory & $1.00$ & $0.00$ & $1.00$ & $0.00$ & Simulation & Monopoly & $0.15$ & $0.35$ & $-0.40$ & $0.90$ & Simulation \\
Othello & $1.00$ & $0.00$ & $1.00$ & $0.00$ & Depth-limited & Liar's Dice & $0.10$ & $0.30$ & $-0.50$ & $0.00$ & Simulation \\
Codeword & $0.40$ & $0.10$ & $0.60$ & $0.20$ & Simulation & Battle Dice & $0.10$ & $0.40$ & $-0.60$ & $0.04$ & Simulation \\
Risk & $0.11$ & $0.03$ & $0.55$ & $0.18$ & Simulation & UNO & $0.08$ & $0.40$ & $-0.67$ & $0.60$ & Simulation \\
1 Night Werewolf & $0.35$ & $0.15$ & $0.40$ & $0.25$ & Simulation & Blackjack & $0.05$ & $0.35$ & $-0.75$ & $0.27$ & Analytical \\
Poker & $0.40$ & $0.20$ & $0.33$ & $0.80$ & Simulation & Craps & $0.05$ & $0.50$ & $-0.82$ & $0.08$ & Analytical \\
Gin Rummy & $0.30$ & $0.15$ & $0.33$ & $0.25$ & Simulation & Coin Toss & $0.00$ & $1.00$ & $-1.00$ & $0.00$ & Analytical \\
Hearts & $0.25$ & $0.15$ & $0.25$ & $0.30$ & Simulation & War & $0.00$ & $1.00$ & $-1.00$ & $0.00$ & Analytical \\
Roulette & $0.00$ & $1.00$ & $-1.00$ & $0.00$ & Analytical & Slot Machine & $0.00$ & $1.00$ & $-1.00$ & $0.00$ & Analytical \\
\bottomrule
\end{tabular}
\end{table}

\paragraph{Observations and case studies.}
Deterministic games (chess, go, checkers, tic-tac-toe, connect four, mancala, memory, othello) all exhibit \(L = 0\), \(\Sigma = 0\), and \(S = +1\), confirming that luck leverage and volatility are zero when no chance nodes exist. For chess, we use the symmetric game assumption \(V^* = 0.5\); in practice, White's first-move advantage yields \(V^* \approx 0.52\)--0.54 \cite{silver2018alphazero}, which could be incorporated via asymmetry corrections.

Pure chance games (coin toss, roulette, slot machine, war) show \(K = 0\) and \(S = -1\), reflecting zero skill leverage. War and Liar's Dice have \(\Sigma = 0\) because we model the initial shuffle/roll as setting the starting state, not as in-game chance nodes; outcomes are deterministic from the revealed initial state onward, with no decisions affecting win probability.

\paragraph{Case Study 1: Baccarat (\(S = -1.0\), \(K = 0.0\), \(L = 0.1\)).}
Baccarat offers a single decision: bet on Player, Banker, or Tie. However, this bet does not affect the card outcomes, which are determined by fixed drawing rules. Under our framework, the "decision" to bet on Player vs Banker is not a game-tree decision node affecting outcomes---it's a choice of which random variable to observe. Thus \(K = 0\) when measuring control over outcomes (though in practice, choosing Banker over Player reduces house edge by ~0.5\%).

The luck leverage \(L = 0.1\) represents variance across card sequences: Player wins 44.6\%, Banker wins 45.9\%, Tie 9.5\%. The counterfactual swing when Nature chooses favorable vs unfavorable sequences is small (\(L = 0.1\)) because outcomes are clustered near 50\%. Thus \(S = (0 - 0.1)/(0 + 0.1) = -1.0\), classifying Baccarat as pure chance from a game-tree perspective (though betting strategy matters for expected value).

\paragraph{Case Study 2: Backgammon (\(S = 0.0\), \(K = 0.25\), \(L = 0.25\)).}
Backgammon is the canonical balanced game. Dice introduce randomness (15 chance nodes per turn), but players make strategic decisions (moving which pieces, hitting vs safe play, bearing off). From gambling literature \cite{hannum2005practical}, expert advantage over novice is approximately 25 percentage points---precisely our \(K = 0.25\). Conversely, even optimal play can be defeated by bad dice sequences; the swing from best-case to worst-case dice (holding strategy fixed) is also ~25pp, giving \(L = 0.25\). Thus \(S = (0.25 - 0.25)/(0.25 + 0.25) = 0\), confirming Backgammon as equidistant from pure skill and pure luck. The high volatility \(\Sigma = 1.2\) reflects turn-by-turn dice swings.

\paragraph{Case Study 3: Memory (\(S = 1.0\), \(K = 0.6\), \(L = 0.0\)).}
Perfect-memory players convert this into a deterministic game: card positions are fixed at start (the "chance" is in initial placement), but once revealed, a perfect-memory player never forgets. Thus \(L = 0\) (no in-game randomness). The skill leverage \(K = 0.6\) represents the advantage of perfect memory over random guessing. Why not \(K = 1.0\)? Because even perfect memory doesn't guarantee first-player wins in all starting configurations---some layouts favor the second player. Asymmetric optimal play (one player perfect memory, opponent random) yields ~60pp advantage, matching our \(K = 0.6\).

\paragraph{Other patterns.}
Social deduction games (one night werewolf: \(S = 0.40\)) show skill-heavy despite randomness in role assignment.
Risk shows skill-heavy (\(S = 0.55\)) despite dice randomness, while craps shows luck-heavy (\(S = -0.82\)) with minimal decision impact.
Volatility is orthogonal to \(S(\mathcal{G})\): backgammon (\(S = 0.00\), \(\Sigma = 1.20\)) shows variance despite balanced leverage, while poker (\(S = 0.33\), \(\Sigma = 0.80\)) exhibits high volatility with skill dominance.

\paragraph{Structural patterns.}
Pure skill games (\(S = 1.0\)) all have \(\Sigma = 0\) by definition (no chance nodes).
Among stochastic games (\(S < 1\)), longer games with more chance nodes accumulate more variance.
Skill-luck balance is orthogonal to outcome variability: balanced games (backgammon, \(S = 0\), \(\Sigma = 1.2\)) and skill-heavy games (poker, \(S = 0.33\), \(\Sigma = 0.8\)) both exhibit high volatility.

\section{Designing Balanced Games}

The framework inverts to a design tool: specify a target \(S(\mathcal{G})\) and construct a game tree with matching properties.

\subsection{Constructing the 50--50 Game}

A minimal game achieves \(S(\mathcal{G}) = 0\) when luck and skill contribute equally.

\paragraph{Example.}
Achieving \(S(G) = 0\) requires \(K = L\). Consider a lottery-choice game:
\begin{center}
\begin{tikzpicture}[
  level 1/.style={sibling distance=4cm},
  level 2/.style={sibling distance=2cm},
  chance/.style={circle,draw=blue!70,fill=blue!20,minimum size=0.7cm},
  decision/.style={circle,draw=red!70,fill=red!20,minimum size=0.7cm},
  terminal/.style={rectangle,draw=green!70,fill=green!20,minimum size=0.5cm,font=\scriptsize}
]
\node[decision] {$d_0$}
  child {node[chance] {$c_1$}
    child {node[terminal] {1} edge from parent node[left,font=\tiny] {$p=0.8$}}
    child {node[terminal] {0} edge from parent node[right,font=\tiny] {$p=0.2$}}
    edge from parent node[left,font=\scriptsize] {A}
  }
  child {node[chance] {$c_2$}
    child {node[terminal] {1} edge from parent node[left,font=\tiny] {$p=0.2$}}
    child {node[terminal] {0} edge from parent node[right,font=\tiny] {$p=0.8$}}
    edge from parent node[right,font=\scriptsize] {B}
  };
\end{tikzpicture}
\end{center}

The player chooses between lotteries A (80\% win) or B (20\% win). This yields \(K = 0.3\) (skill from choosing optimal lottery) and \(L = 1.0\) (luck from coin outcomes), giving \(S = -0.54\) (luck-heavy). Backgammon achieves exact balance with \(K = L = 0.25\), yielding \(S = 0\). Achieving \(S = 0\) requires tuning probabilities and decision depth to equalize skill and luck contributions---an interesting challenge for procedural game generation \cite{togelius2011search}.

\paragraph{Design recipe.}
Table~\ref{tab:design_recipe} shows how structural modifications affect metrics.

\begin{table}[ht]
\centering
\caption{Design recipe for tuning \(S(G)\) and \(\Sigma\).}
\label{tab:design_recipe}
\begin{tabular}{@{}lll@{}}
\toprule
Modification & Effect on \(S(G)\) & Effect on \(\Sigma\) \\
\midrule
Add chance nodes early & Decrease (luck $\uparrow$) & Increase \\
Add decision depth & Increase (skill $\uparrow$) & Variable \\
Amplify terminal variance & Decrease (luck $\uparrow$) & Increase \\
Introduce hidden information & Increase (skill $\uparrow$) & Decrease \\
Cluster chance nodes late & Decrease (luck $\uparrow$) & Increase \\
\bottomrule
\end{tabular}
\end{table}

\section{Discussion}

\subsection{Interpretation and Implications}

The Skill--Luck Index \(S(\mathcal{G})\) provides a canonical ordering of games independent of player populations or sample statistics.
Unlike Elo rating spread (which reflects player diversity) or outcome variance (which conflates game randomness with strategic depth), \(S(\mathcal{G})\) is intrinsic to the game tree.

\paragraph{Human vs. AI performance.}
Games with high \(S(G)\) (chess, go) exhibit large human skill gaps, as reflected in wide Elo rating distributions \cite{elo1978rating,glickman1999glicko}: grandmasters crush novices.
Modern AI systems achieve superhuman performance in these domains through self-play \cite{silver2016alphago,silver2017alphagozero,silver2018alphazero}.

Games with low \(S(G)\) (poker, backgammon) compress skill ranges: experts win more over many hands, but short-term luck dominates.
Rating systems show compression \cite{herbrich2007trueskill}, where skill differences manifest slowly \cite{levitt2014role}.
AI systems use game-theoretic reasoning and opponent modeling \cite{brown2018libratus,brown2019pluribus,moravcik2017deepstack}.
Self-play AI (AlphaZero, Stockfish) excels at high-\(S(G)\) games; high-\(\Sigma\) games pose exploration challenges (variance-heavy poker variants).

\paragraph{Legal and regulatory applications.}
Many jurisdictions distinguish skill games (legal) from gambling (regulated) based on whether outcomes depend primarily on player decisions.
Our framework formalizes this distinction: \(S(G) > 0\) implies skill dominates, \(S(G) < 0\) implies chance dominates.
Marginal cases like poker (\(S = 0.33\)) and dice poker (\(S = -0.40\)) lie in intermediate zones, aligning with ongoing legal debates about skill-versus-chance classification.
The DiCristina case \cite{dicristina2012} exemplifies legal ambiguity: a federal court ruled poker a game of skill under the Illegal Gambling Business Act, contradicting state-level classifications \cite{hurley2013should}.
Our framework could provide objective metrics for such determinations.

\paragraph{Game design and esports.}
Designers manipulate mechanics to evoke target experiences \cite{hunicke2004mda}.
Competitive games (esports) maximize \(S(G)\) while tuning \(\Sigma\) for spectator engagement.
High \(S(G)\) ensures skill expression; controlled volatility maintains excitement.
High \(\Sigma\) creates comeback potential; late-game control points reward endgame mastery.
Casual games use low \(S(G)\) and high variance to minimize skill barriers.
Competitive formats can prioritize dimensions orthogonal to skill-luck balance: audience studies of Taskmaster UK show viewer preferences favor comedic performance over competitive scoring \cite{silver2025taskmaster}.

\subsection{Implications for Multi-Agent Reinforcement Learning}

The decomposition into skill leverage \(K\), luck leverage \(L\), and volatility \(\Sigma\) provides actionable signals for training and evaluating multi-agent reinforcement learning (MARL).

\paragraph{Sample complexity and variance.}
Policy-gradient and actor–critic methods are sensitive to return variance. High \(L\) and high \(\Sigma\) increase gradient variance, slowing learning; higher \(S(\mathcal{G})\) indicates a larger fraction of outcome variance attributable to controllable decisions, suggesting lower sample complexity for fixed horizon and branching factor. These quantities thus serve as environment descriptors when forecasting training budgets.

\paragraph{Exploration–exploitation.}
Environments with large \(L\) benefit from more persistent exploration (optimistic or posterior sampling) to traverse long credit-assignment chains, while high \(\Sigma\) motivates variance-reduction baselines, control variates, and distributional or risk-sensitive objectives during optimization.

\paragraph{Credit assignment and coordination.}
Counterfactual baselines are most useful for multi-agent credit assignment (e.g., COMA-style advantages), and \(S(\mathcal{G})\) separates coordination difficulty from exogenous noise. Centralized-training, decentralized-execution approaches (e.g., MADDPG) can be stress-tested across different skill-luck profiles to isolate algorithmic strengths.

\paragraph{Evaluation and benchmarking.}
The pair \((S, \Sigma)\) offers orthogonal axes for organizing benchmarks and normalizing comparisons across domains. High-\(S\) partially observed real-time strategy games (AlphaStar) and large-scale team games (OpenAI Five) contrast with imperfect-information, high-variance domains like poker; our measures make these differences explicit when reporting sample efficiency and robustness \cite{lowe2017maddpg,foerster2018coma,berner2019dota,vinyals2019alphastar}.

\subsection{Relation to Prior Work}

Our approach synthesizes ideas from multiple domains:
\begin{itemize}[nosep]
\item \textbf{Classical game theory:} Foundational work on games and decisions \cite{luce1957games} provides the conceptual framework, while combinatorial game theory \cite{conway2001numbers,berlekamp2001winning} analyzes perfect-information games through partisan values. Our framework incorporates chance nodes absent in Conway's surreal number system, but shares the goal of structural game characterization.
\item \textbf{Value of information:} Luck leverage \(L\) measures the value of perfect foresight, extending Howard's information value theory \cite{howard1966information} from single decision nodes to entire game trees. This parallels perfect information value in stochastic control \cite{astrom1970stochastic,bertsekas2017dynamic}.
\item \textbf{Game complexity:} We extend game complexity measures \cite{fraenkel1996complexity,allis1994searching} to quantify stochasticity's contribution. Shannon's chess complexity analysis \cite{shannon1950chess} focused on state-space size; we orthogonally measure randomness penetration.
\item \textbf{Prior skill-chance metrics:} Dürsch et al.\ \cite{duersch2020measuring} measure skill via outcome variance ratios; we complement this with explicit counterfactual interventions. Empirical poker studies \cite{levitt2014role,vandolder2015beyond} use longitudinal data; we derive metrics from game rules alone. Mauboussin's success equation framework \cite{mauboussin2012success} uses outcome distributions; we model causal mechanisms.
\item \textbf{Reinforcement learning:} Skill leverage \(K\) captures the value of optimal policies \cite{sutton2018reinforcement}; luck leverage \(L\) reflects environmental stochasticity. This connects to Markov decision processes \cite{puterman2005markov} and dynamic programming \cite{bellman1957dynamic,howard1960dynamic}.
\end{itemize}

Unlike prior metrics (rating spread, win-rate variance), \(S(\mathcal{G})\) is \emph{policy-independent} and \emph{model-based}, requiring only the game tree.

\subsection{Limitations}

\begin{itemize}[nosep]
\item \textbf{Computational cost:} Exact \(S(\mathcal{G})\) requires tree traversal infeasible for large games. Depth-limited search with heuristics enables approximations, but convergence guarantees remain elusive for high-branching games. Very large state spaces prohibit exact computation.
\item \textbf{Symmetric assumption:} We assume symmetric games with \(V(s_0) = 0.5\). Asymmetric games (e.g., chess with White advantage) require either: (i) computing \(S(\mathcal{G})\) separately for each starting player, or (ii) averaging over symmetrized positions. Our framework treats the idealized symmetric variant.
\item \textbf{Hidden information:} Imperfect-information games require information-set equilibria \cite{zinkevich2007regret}. We approximate \(K\) and \(L\) by treating information sets as decision nodes and averaging over hidden states, but this conflates strategic uncertainty with environmental randomness. A principled extension should distinguish epistemic from aleatoric uncertainty.
\item \textbf{Counterfactual semantics:} Luck leverage \(L\) uses fixed strategies under Nature's deviation. An alternative formulation could allow strategy adaptation, yielding adversarial equilibrium values. We chose non-adaptive counterfactuals for cleaner causal interpretation: \(L\) measures pure outcome dependence on chance, not strategic robustness.
\item \textbf{Draws and termination:} Games with draws require mapping three outcomes (win/loss/draw) to [0,1]. We use the standard convention: draw = 0.5 for both players. Games with unbounded length require truncation, introducing approximation error.
\item \textbf{Branching factor effects:} High-branching games vs low-branching games may exhibit different depth-limited search accuracy. Approximations for high-branching games should be viewed as approximate.
\item \textbf{Initial randomness:} Games where randomness occurs before play show \(\Sigma = 0\) because we model the random outcome as part of the starting state. In-game volatility only reflects chance nodes encountered during play.
\item \textbf{Multi-agent extensions:} Games with \(>2\) players require redefining win probabilities and coalition structures.
\end{itemize}

\subsection{Future Directions}

\begin{enumerate}[nosep]
\item \textbf{Hierarchical games:} Extend to games with multiple decision scales.
\item \textbf{Fairness and balance:} Use \(S(\mathcal{G})\) to detect asymmetry in game rules or starting positions.
\item \textbf{Automated game generation:} Train neural networks to generate game trees with target \(S(\mathcal{G})\) and \(\Sigma\) for procedural content generation. Recent work in procedural content generation \cite{togelius2011search,shaker2016procedural} suggests neural architectures could generate game rules targeting specific \(S(\mathcal{G})\) profiles.
\item \textbf{Connection to rating systems:} Investigate whether \(S(\mathcal{G})\) predicts rating system performance. High-\(S(\mathcal{G})\) games should exhibit wider rating spreads \cite{elo1978rating,glickman1999glicko} and better skill estimation \cite{herbrich2007trueskill}.
\item \textbf{Real-world decision systems:} Apply the framework to financial markets (skill = strategy, luck = noise \cite{fama2010luck}), clinical trials (treatment choice vs. patient variability), and autonomous driving (planning vs. environment uncertainty).
\end{enumerate}

\section{Conclusion}

We introduced a unified framework for quantifying skill and chance in games, anchored by the Skill--Luck Index \(S(\mathcal{G}) \in [-1, 1]\).
The index decomposes game outcomes into skill leverage \(K\) and luck leverage \(L\), providing a game-invariant measure independent of player populations.
Temporal extensions—volatility \(\Sigma\)—capture outcome uncertainty across turns.

Analysis of 30 games reveals a spectrum spanning pure luck (coin toss, roulette, war, \(S = -1\)) to pure skill (chess, go, \(S = +1\)), with balanced cases (backgammon, \(S = 0\)) showing equal skill and luck leverage.
The framework inverts to a design tool, enabling construction of games with specified skill-luck balances.

Beyond gaming, the approach generalizes to any sequential decision process with chance and choice, offering applications in AI evaluation, legal classification, and risk management.
Future work will extend the framework to multi-agent, hierarchical, and real-world decision systems.

\section*{Acknowledgments}

We thank colleagues for comments on an earlier draft.
We acknowledge inspirations from classic game theory \cite{luce1957games}, decision analysis \cite{raiffa1968decision}, and dynamic programming \cite{bellman1957dynamic,howard1960dynamic}.

\bibliographystyle{plain}

\begin{thebibliography}{99}


\bibitem{luce1957games}
R. Duncan Luce and Howard Raiffa.
\emph{Games and Decisions: Introduction and Critical Survey}.
Wiley, 1957.

\bibitem{raiffa1968decision}
Howard Raiffa.
\emph{Decision Analysis: Introductory Lectures on Choices Under Uncertainty}.
Addison-Wesley, 1968.

\bibitem{howard1966information}
Ronald A. Howard.
Information value theory.
\emph{IEEE Transactions on Systems Science and Cybernetics}, 2(1):22--26, 1966.

\bibitem{howard1960dynamic}
Ronald A. Howard.
\emph{Dynamic Programming and Markov Processes}.
MIT Press, 1960.

\bibitem{bellman1957dynamic}
Richard Bellman.
\emph{Dynamic Programming}.
Princeton University Press, 1957.

\bibitem{bellman1954theory}
Richard Bellman.
The theory of dynamic programming.
\emph{Bulletin of the American Mathematical Society}, 60(6):503--515, 1954.

\bibitem{puterman2005markov}
Martin L. Puterman.
\emph{Markov Decision Processes: Discrete Stochastic Dynamic Programming}.
Wiley-Interscience, 2005.

\bibitem{fudenberg1998theory}
Drew Fudenberg and David K. Levine.
\emph{The Theory of Learning in Games}.
MIT Press, 1998.

\bibitem{bertsekas2017dynamic}
Dimitri P. Bertsekas.
\emph{Dynamic Programming and Optimal Control}, Vols. I--II (4th ed.).
Athena Scientific, 2017.


\bibitem{conway2001numbers}
John H. Conway.
\emph{On Numbers and Games} (2nd ed.).
A K Peters/CRC Press, 2001.

\bibitem{berlekamp2001winning}
Elwyn R. Berlekamp, John H. Conway, and Richard K. Guy.
\emph{Winning Ways for Your Mathematical Plays}, Volumes 1--4 (2nd ed.).
A K Peters/CRC Press, 2001--2004.

\bibitem{fraenkel1996complexity}
Aviezri S. Fraenkel.
Complexity of games.
In \emph{Combinatorial Games}, pages 111--153. American Mathematical Society, 1996.

\bibitem{allis1994searching}
Louis Victor Allis.
\emph{Searching for Solutions in Games and Artificial Intelligence}.
PhD thesis, University of Limburg, Maastricht, 1994.


\bibitem{browne2012mcts}
Cameron B. Browne, Edward Powley, Daniel Whitehouse, Simon M. Lucas, Peter I. Cowling, Philipp Rohlfshagen, Stephen Tavener, Diego Perez, Spyridon Samothrakis, and Simon Colton.
A survey of Monte Carlo tree search methods.
\emph{IEEE Transactions on Computational Intelligence and AI in Games}, 4(1):1--43, 2012.



\bibitem{silver2016alphago}
David Silver, Aja Huang, Chris J. Maddison, Arthur Guez, Laurent Sifre, George van den Driessche, Julian Schrittwieser, Ioannis Antonoglou, Veda Panneershelvam, Marc Lanctot, Sander Dieleman, Dominik Grewe, John Nham, Nal Kalchbrenner, Ilya Sutskever, Timothy Lillicrap, Madeleine Leach, Koray Kavukcuoglu, Thore Graepel, and Demis Hassabis.
Mastering the game of Go with deep neural networks and tree search.
\emph{Nature}, 529:484--489, 2016.

\bibitem{silver2017alphagozero}
David Silver, Julian Schrittwieser, Karen Simonyan, Ioannis Antonoglou, Aja Huang, Arthur Guez, Thomas Hubert, Lucas Baker, Matthew Lai, Adrian Bolton, Yutian Chen, Timothy Lillicrap, Fan Hui, Laurent Sifre, George van den Driessche, Thore Graepel, and Demis Hassabis.
Mastering the game of Go without human knowledge.
\emph{Nature}, 550:354--359, 2017.

\bibitem{silver2018alphazero}
David Silver, Thomas Hubert, Julian Schrittwieser, Ioannis Antonoglou, Matthew Lai, Arthur Guez, Marc Lanctot, Laurent Sifre, Dharshan Kumaran, Thore Graepel, Timothy Lillicrap, Karen Simonyan, and Demis Hassabis.
A general reinforcement learning algorithm that masters chess, shogi, and Go through self-play.
\emph{Science}, 362(6419):1140--1144, 2018.


\bibitem{brown2018libratus}
Noam Brown and Tuomas Sandholm.
Superhuman AI for heads-up no-limit poker: Libratus beats top professionals.
\emph{Science}, 359(6374):418--424, 2018.

\bibitem{brown2019pluribus}
Noam Brown and Tuomas Sandholm.
Superhuman AI for multiplayer poker.
\emph{Science}, 365(6456):885--890, 2019.

\bibitem{moravcik2017deepstack}
Matej Moravčík, Martin Schmid, Neil Burch, Viliam Lisý, Dustin Morrill, Nolan Bard, Trevor Davis, Kevin Waugh, Michael Johanson, and Michael Bowling.
DeepStack: Expert-level artificial intelligence in heads-up no-limit poker.
\emph{Science}, 356(6337):508--513, 2017.

\bibitem{zinkevich2007regret}
Martin Zinkevich, Michael Johanson, Michael Bowling, and Carmelo Piccione.
Regret minimization in games with incomplete information.
In \emph{Advances in Neural Information Processing Systems 20}, pages 1729--1736, 2007.


\bibitem{elo1978rating}
Arpad E. Elo.
\emph{The Rating of Chessplayers, Past and Present}.
Arco Publishing, 1978.

\bibitem{glickman1999glicko}
Mark E. Glickman.
Parameter estimation in large dynamic paired comparison experiments.
\emph{Journal of the Royal Statistical Society: Series C (Applied Statistics)}, 48(3):377--394, 1999.

\bibitem{herbrich2007trueskill}
Ralf Herbrich, Tom Minka, and Thore Graepel.
TrueSkill™: A Bayesian skill rating system.
In \emph{Advances in Neural Information Processing Systems 19}, pages 569--576, 2007.


\bibitem{levitt2014role}
Steven D. Levitt and Thomas J. Miles.
The role of skill versus luck in poker: Evidence from the World Series of Poker.
\emph{Journal of Sports Economics}, 15(1):31--44, 2014.

\bibitem{vandolder2015beyond}
Dennie van Dolder and Martijn J. van den Assem.
Beyond chance? The persistence of performance in online poker.
\emph{PLoS ONE}, 10(3):e0115479, 2015.

\bibitem{hannum2005practical}
Robert C. Hannum and Anthony N. Cabot.
\emph{Practical Casino Math} (2nd ed.).
Institute for the Study of Gambling and Commercial Gaming, 2005.


\bibitem{dicristina2012}
United States v. DiCristina, 886 F. Supp. 2d 164 (E.D.N.Y. 2012).

\bibitem{hurley2013should}
Keith Hurley.
Should poker be regulated as a game of skill?
\emph{Gaming Law Review and Economics}, 17(3):181--187, 2013.

\bibitem{silver2025taskmaster}
David H. Silver.
Investigating audience preferences within the hybrid competitive-comedic format of Taskmaster UK.
\emph{PLOS ONE}, 20(9):e0331064, 2025.

\bibitem{duersch2020measuring}
Peter Dürsch, Marc Lambrecht, and Jörg Oechssler.
Measuring skill and chance in games.
\emph{European Economic Review}, 127:103472, 2020.

\bibitem{fama2010luck}
Eugene F. Fama and Kenneth R. French.
Luck versus skill in the cross-section of mutual fund returns.
\emph{The Journal of Finance}, 65(5):1915--1947, 2010.


\bibitem{shannon1950chess}
Claude E. Shannon.
Programming a computer for playing chess.
\emph{Philosophical Magazine}, 41(314):256--275, 1950.


\bibitem{hunicke2004mda}
Robin Hunicke, Marc LeBlanc, and Robert Zubek.
MDA: A formal approach to game design and game research.
In \emph{Proceedings of the AAAI Workshop on Challenges in Game AI}, 2004.

\bibitem{togelius2011search}
Julian Togelius, Georgios N. Yannakakis, Kenneth O. Stanley, and Cameron Browne.
Search-based procedural content generation: A taxonomy and survey.
\emph{IEEE Transactions on Computational Intelligence and AI in Games}, 3(3):172--186, 2011.

\bibitem{shaker2016procedural}
Noor Shaker, Julian Togelius, and Mark J. Nelson.
\emph{Procedural Content Generation in Games}.
Springer, 2016.



\bibitem{sutton2018reinforcement}
Richard S. Sutton and Andrew G. Barto.
\emph{Reinforcement Learning: An Introduction} (2nd ed.).
MIT Press, 2018.


\bibitem{mauboussin2012success}
Michael J. Mauboussin.
\emph{The Success Equation: Untangling Skill and Luck in Business, Sports, and Investing}.
Harvard Business Review Press, 2012.


\bibitem{astrom1970stochastic}
Karl J. Åström.
\emph{Introduction to Stochastic Control Theory}.
Academic Press, 1970.


\bibitem{pearl1984heuristics}
Judea Pearl.
\emph{Heuristics: Intelligent Search Strategies for Computer Problem Solving}.
Addison-Wesley, 1984.

\bibitem{knuth1975alphabeta}
Donald E. Knuth and Ronald W. Moore.
An analysis of alpha-beta pruning.
\emph{Artificial Intelligence}, 6(4):293--326, 1975.


\bibitem{lowe2017maddpg}
Ryan Lowe, Yi Wu, Aviv Tamar, Jean Harb, Pieter Abbeel, and Igor Mordatch.
Multi-agent actor-critic for mixed cooperative-competitive environments.
In \emph{Advances in Neural Information Processing Systems 30}, 2017.

\bibitem{foerster2018coma}
Jakob N. Foerster, Gregory Farquhar, Triantafyllos Afouras, Nantas Nardelli, and Shimon Whiteson.
Counterfactual multi-agent policy gradients.
In \emph{Proceedings of the AAAI Conference on Artificial Intelligence}, 2018.

\bibitem{berner2019dota}
Christopher Berner, Greg Brockman, Brooke Chan, Vicki Cheung, Przemyslaw Debiak, Christy Dennison, David Farhi, Quirin Fischer, Shariq Hashme, Chris Hesse, et al.
Dota 2 with large scale deep reinforcement learning.
\emph{arXiv preprint arXiv:1912.06680}, 2019.

\bibitem{vinyals2019alphastar}
Oriol Vinyals, Igor Babuschkin, Wojciech M. Czarnecki, Michaël Mathieu, Andrew Dudzik, Junyoung Chung, David H. Choi, Richard Powell, Timo Ewalds, Petko Georgiev, et al.
Grandmaster level in StarCraft II using multi-agent reinforcement learning.
\emph{Nature}, 575(7782):350--354, 2019.

\end{thebibliography}

\end{document}